\documentclass[journal]{IEEEtran}
\usepackage{cite}
\usepackage{times}
\usepackage{epsfig}
\usepackage{graphicx}
\usepackage{amsmath}
\usepackage{amssymb}
\usepackage{subcaption}

\usepackage{array}
\usepackage{mathrsfs}
\usepackage{booktabs}
\usepackage[pagebackref=true,breaklinks=true,letterpaper=true,colorlinks,bookmarks=false]{hyperref}
\usepackage{amsmath}
\usepackage{amssymb}
\usepackage{algorithmic}
\usepackage{color}

\usepackage{amsmath}
\usepackage{amssymb}

\usepackage{color}
\usepackage{epsfig}
\usepackage{epstopdf}
\usepackage{algorithm}
\usepackage{algorithmic}
\usepackage{multirow}
\usepackage{comment}
\usepackage{diagbox}

\usepackage[switch]{lineno}

\begin{document}
	
%
\title{Learning Dynamical Human-Joint Affinity for 3D Pose Estimation in Videos}
%
%
%

\author{Junhao Zhang,
	    Yali Wang,
        Zhipeng Zhou,
        Tianyu Luan,
        Zhe Wang
        and~Yu~Qiao,~\IEEEmembership{Senior Member,~IEEE}
\thanks{
J. Zhang, Y. Wang and Z. Zhou are equally-contributed authors.

J. Zhang, Y. Wang, Z. Zhou, T. Luan, and Y. Qiao are with ShenZhen Key Lab of Computer Vision and Pattern Recognition, Shenzhen Institute of Advanced Technology, Chinese Academy of Sciences. Z. Wang is with the University of California, Irvine. Y. Qiao is also with Shanghai AI Laboratory, Shanghai, China.

Y. Qiao is \emph{the corresponding author}. Email: yu.qiao@siat.ac.cn
}

}

%
%

\markboth{IEEE TRANSACTIONS ON IMAGE PROCESSING}%
{Shell \MakeLowercase{\textit{Zhang et al.}}: IEEE TRANSACTIONS ON IMAGE PROCESSING}

%



\maketitle

\begin{abstract}
Graph Convolution Network (GCN) has been successfully used for 3D human pose estimation in videos. %
However,
it is often built on the fixed human-joint affinity,
according to human skeleton.
This may reduce adaptation capacity of GCN to tackle complex spatio-temporal pose variations in videos.
To alleviate this problem,
we propose a novel Dynamical Graph Network (DG-Net),
which can dynamically identify human-joint affinity,
and estimate 3D pose by adaptively learning spatial/temporal joint relations from videos. %
Different from traditional graph convolution,
we introduce Dynamical Spatial/Temporal Graph convolution (DSG/DTG) to discover spatial/temporal human-joint affinity for each video exemplar,
depending on spatial distance/temporal movement similarity between human joints in this video.
Hence,
they can effectively understand which joints are spatially closer and/or have consistent motion,
for reducing depth ambiguity and/or motion uncertainty when lifting 2D pose to 3D pose.
We conduct extensive experiments on three popular benchmarks,
e.g.,
Human3.6M,
HumanEva-I,
and
MPI-INF-3DHP,
where
DG-Net outperforms a number of recent SOTA approaches with fewer input frames and model size.
\end{abstract}


%
\IEEEpeerreviewmaketitle

\begin{figure*}[t]
\centering
\includegraphics[width=1\textwidth]{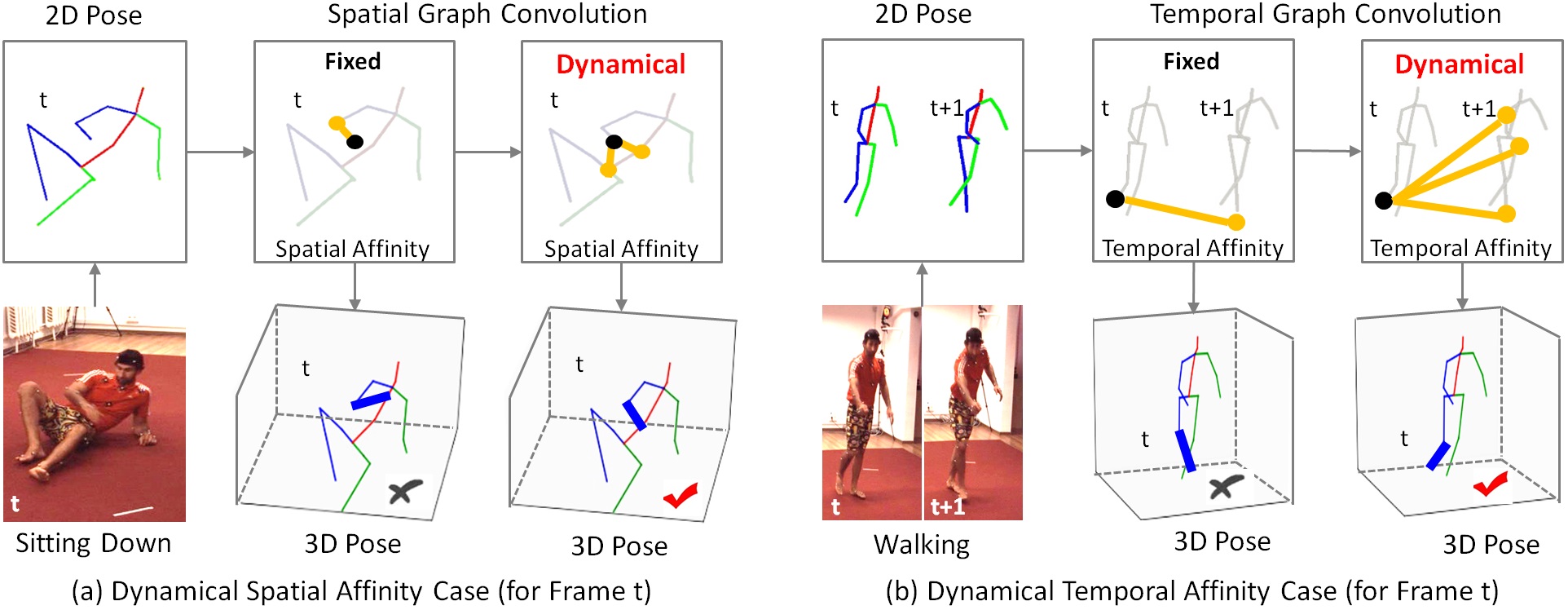}
\caption{Our motivations.
The fixed human-joint affinity often limits adaptation capacity of spatial/temporal graph convolution to tackle complex pose variations in videos.
To alleviate such difficulty,
the graph affinity should depend on the specific human pose in the video.
To achieve this goal,
we propose to learn spatial/temporal human-joint affinity dynamically,
according to spatial/temporal similarity between joints.
For example,
\textit{wrist} is spatially closer to \textit{hip} and \textit{pelvis} in the \textit{SittingDown} case of Fig.\ref{motivation}(a).
Hence,
we should further enhance such spatial affinity to estimate \textit{wrist} at Frame $t$.
Similarly,
\textit{ankle} keeps the consistent movement with \textit{wrist} and \textit{shoulder} in the \textit{Walking} case.
Hence,
we should further enhance such temporal affinity to estimate \textit{ankle} at Frame $t$.
More explanations can be found in the introduction.
}
\label{motivation}
\vspace{-0.3cm}
\end{figure*}

\section{Introduction}
%
%
%
%
\IEEEPARstart{T}{his} paper focuses on the problem of 3D pose estimation in videos.
It is an important computer vision task,
due to its wide applications in
video surveillance,
virtual reality,
etc.
With rapid development of deep learning,
this task has achieved remarkable progresses \cite{Lin_2017_CVPR,eccv2018temporal,pavllo:videopose3d:2019}.
In particular,
recent studies \cite{semanticsgcn,Ci_2019_ICCV,Cai_2019_ICCV}  have shown that,
graph convolution networks can effectively boost 3D pose estimation in videos,
by learning spatio-temporal relations from 2D pose.
However,
spatial/temporal graph convolutions in these models are built upon the fixed human-joint affinity that is defined by human skeleton.
As a result,
it is often limited to tackle depth ambiguity and/or motion uncertainty of human joints in videos,
without taking dynamical pose variations into account.
We use two examples in Fig.\ref{motivation} to illustrate this problem.

First,
we discuss spatial graph convolution by a \textit{SittingDown} case at Frame $t$.
As shown in Fig.\ref{motivation}(a),
\textit{wrist} is connected with \textit{elbow} in the human skeleton.
Based on such fixed spatial affinity,
spatial graph convolution leverages \textit{elbow} as the main context to update \textit{wrist}.
But,
\textit{elbow} stays at a higher position that is relatively far from \textit{wrist} in this \textit{SittingDown} case.
Consequently,
when integrating context from \textit{elbow},
spatial graph convolution would lift \textit{wrist} to an unsatisfactory position that is higher than its ground truth.
In fact,
\textit{wrist} is spatially closer to \textit{hip} and \textit{pelvis} in this \textit{SittingDown} case.
Hence,
it is necessary to further exploit such spatial affinity and learn context from these joints for estimating 3D position of \textit{wrist}.

Second,
we discuss temporal graph convolution by a \textit{Walking} case from Frame $t$ to $t+1$.
According to the fixed temporal affinity in Fig.\ref{motivation}(b),
\textit{ankle} at Frame $t$ is connected with the same joint \textit{ankle} at Frame $t+1$.
In this case,
temporal graph convolution leverages \textit{ankle} at Frame $t+1$ as temporal context for estimating \textit{ankle} at Frame $t$.
But,
\textit{ankle} at Frame $t+1$ is moving forward.
Only integrating such context would mislead temporal graph convolution to estimate \textit{ankle} at Frame $t$ to be forward.
To capture how \textit{ankle} moves correctly,
it would be better to use other joints with similar motion as contextual guidance.
In fact,
in this \textit{Walking} case,
\textit{ankle} keeps the consistent movement with \textit{wrist} and \textit{shoulder} from Frame $t$ to Frame $t+1$.
To reduce motion uncertainty,
it is necessary to further enhance such temporal affinity,
and integrate contexts from \textit{wrist} and \textit{shoulder} at Frame $t+1$ for estimating 3D position of \textit{ankle} at $t$.

Both cases show that,
the fixed human-joint affinity often lacks adaptation capacity to describe the personalized pose dynamics in videos.
To alleviate this difficulty,
we propose a novel Dynamical Graph Network (DG-Net),
which can dynamically adjust spatial/temporal human-joint affinity in videos,
and adaptively integrate joint-dependent context for accurate 3D pose estimation.
More specifically,
we first design two dynamical graph convolution operations,
i.e.,
Dynamical Spatial Graph (DSG) and Dynamical Temporal Graph (DTG) convolution.
DSG/DTG can dynamically construct human-joint affinity,
according to similarity of spatial distance/temporal movement between joints.
Hence,
they can capture robust human-joint relations to tackle pose variations in different videos.
Second,
we embed DSG and DTG into a dynamical graph convolution block,
and cascade a number of such blocks progressively to build up DG-Net.
Via adding 3D pose supervision on individual blocks and fusion of different blocks,
our DG-Net can regularize the cooperative power of spatial and temporal joint relations in each block,
and integrate complementary features in different blocks to boost 3D pose estimation in videos.
Finally,
we perform extensive experiments on widely-used human pose benchmarks,
i.e.,
Human3.6M,
HumanEva-I,
and
MPI-INF-3DHP.
Our DG-Net outperforms a number of recent SOTA approaches.

\section{Related Work}

\textbf{2D Pose Estimation}.
2D human pose estimation has been well studied in the decades.
Traditionally,
it is addressed by hand-craft features and pictorial structures \cite{dantone2013human}.
With fast development of deep learning,
2D pose estimation have achieved remarkable successes by using Convolutional Neural Networks (CNNs) \cite{wei2016convolutional,toshev2014deeppose,cpn,hourglass}.
For example,
\cite{toshev2014deeppose} firstly introduces an end-to-end CNN to solve this problem by regressing human pose.
\cite{wei2016convolutional} designs a multi-stage architecture which estimates human pose in a coarse-to-fine manner.
\cite{hourglass} proposes a hourglass network to utilize multi-scale features to capture human pose in the images.
\cite{cpn} develops a cascaded pyramid network to address pose estimation from simple to hard keypoints.
Lately,
\cite{qiu2020dgcn} utilizes dynamical graph to boost 2D pose estimation in the single image.
However,
affinity matrix of \cite{qiu2020dgcn} is actually fixed and same for all the persons in each training iteration.
It is called as a dynamical matrix,
since it is changed over training iterations according to Bernoulli distribution of a precomputed soft matrix.
On the contrary,
our affinity matrix is dynamical,
since it adaptively finds contextual joints to describe spatio-temporal pose relation of each human exemplar in the video.
Additionally,
we work on 3D pose estimation in videos,
while
\cite{qiu2020dgcn} focuses on 2D pose estimation in still images.

\textbf{3D Pose Estimation}.
3D pose estimation has been mainly
driven by deep neural networks. Currently, most approaches can be categorized into two frameworks.
One framework is to directly estimate 3D joint locations from images \cite{mono_3dhp2017,volumetric,integral,ordinal}.
A popular solution in this framework is based on the volumetric representation of human pose \cite{volumetric}.
However,
this method requires the non-differentiable post-processing step and introduces the quantization error \cite{integral}.
To alleviate such difficulty,
several extensions have been proposed by
integral regression \cite{integral},
ordinal depth supervision \cite{ordinal},
etc.
Another framework is to first estimate 2D pose and then lift 2D pose to 3D pose \cite{simple,semanticsgcn,Ci_2019_ICCV,Cai_2019_ICCV,lin2019trajectory,wang2020motion},
due to the fast development of 2D pose estimation \cite{alej2016stacked,cpn}.
For instance,
\cite{simple} introduces a simple but effective residual fully connected network to regress 3D pose from 2D pose.
 \cite{distance} formulates the problem as a 2D-to-3D distance matrix regression,
which is often robust to missing observations.
\cite{eccv2018temporal,Lee_2018_ECCV,pavllo:videopose3d:2019} encode the temporal information in their lifting network,
in order to tackle video 3D pose estimation with temporal smoothness.
However,
these approaches are often limited to learn distinct spatio-temporal relations between human joints in different actions,
i.e.,
an important clue of pose ambiguity reduction.
Alternatively,
our DG-Net can capture rich relations by learning human-joint affinity dynamically,
which effectively boosts 3D pose estimation in videos.


\begin{figure*}[t]
\centering
\includegraphics[width=1\textwidth]{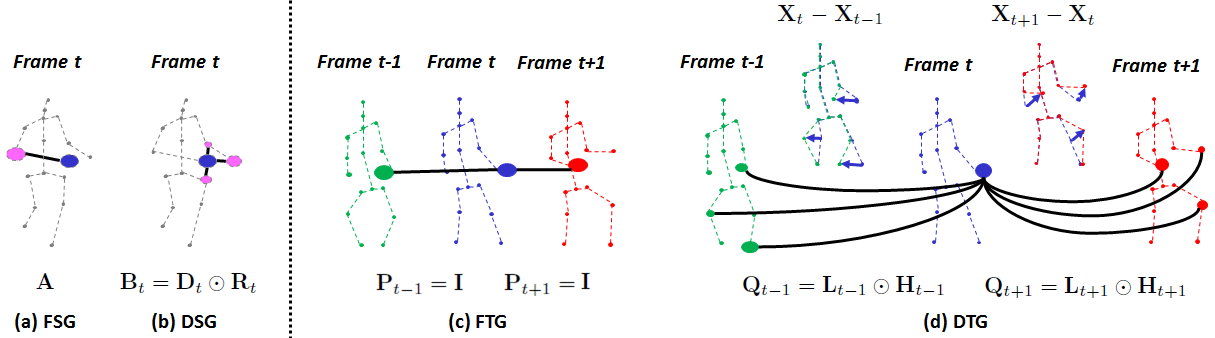}
\caption{Graph Illustration.
(a) Fixed Spatial Graph (FSG).
Graph affinity $\mathbf{A}$ is constructed according to fixed spatial structure of human skeleton.
(b) Dynamical Spatial Graph (DSG).
Graph affinity $\mathbf{B}_{t}$ is constructed by finding connections of spatial neighbors $\mathbf{D}_{t}$ and weighting importance of these neighbors $\mathbf{R}_{t}$.
(c) Fixed Temporal Graph (FTG).
Forward and backward graph affinity $\mathbf{P}_{t+1}$, $\mathbf{P}_{t-1}$ are constructed according to fixed temporal structure of human skeleton.
(d) Dynamical Temporal Graph (DTG).
Forward and backward graph affinity $\mathbf{Q}_{t+1}$, $\mathbf{Q}_{t-1}$ are constructed by finding connections of temporal neighbors $\mathbf{L}_{t+1}$, $\mathbf{L}_{t-1}$ and weighting importance of these neighbors $\mathbf{H}_{t+1}$, $\mathbf{H}_{t-1}$.
The black line denotes the connection between joints.
The size of circle in DSG and DTG refers to the importance of connected neighbors.
More details can be found in Section \ref{Method}.}
\label{model}
\vspace{-0.3cm}
\end{figure*}

\textbf{Graph Convolution Networks}.
Graph Convolution Network (GCN) is widely used to model graph-structured data \cite{HenaffarXiv2015,LiICLR2016,Bronstein2017,maxgcn,PetarICLR18}.
It has been adopted for skeleton-based action recognition \cite{stgcn} and group activity recognition \cite{wu2019learning}.
Different from these works,
we tackle 3D pose estimation with depth ambiguity and/or motion uncertainty.
Recently,
some GCN approaches has been introduced to lift 2D input pose for 3D pose estimation \cite{semanticsgcn,dynamicgraph_gesture,Ci_2019_ICCV,Cai_2019_ICCV}.
For instance,
\cite{semanticsgcn} introduces a semantic GCN,
which learns channel-wise edge weights for enhancing relations between joints.
\cite{Cai_2019_ICCV} proposes a local-to-global GCN network,
in order to learn multi-scale pose relations for 3D estimation.
However,
all these approaches are limited for learning joint relations,
based on the following reasons.
First,
\cite{stgcn,semanticsgcn,Cai_2019_ICCV} are built upon a fixed spatial/temporal graph affinity,
i.e.,
the connections between joints are based on the fixed human skeleton.
Hence,
they are limited to capture complex pose variations in videos.
Second,
\cite{wu2019learning} utilizes graph attention to learn the relation among all the actors,
where
all the actors are fully connected in the graph affinity.
However,
this design may not be suitable for human joints,
i.e.,
every unit only contains the fully-connected graph in the global manner,
which often brings the noisy joint relations when learning local affinity in the pose estimation.
Third,
\cite{Shi_2019_CVPR} is the combination of two cases above.
It utilizes a graph affinity that contains three terms,
i.e.
skeleton term,
data-driven term,
and
attention term.
The skeleton and data-driven terms refer to the first case of the fixed affinity,
while
the attention term refers to the second case of the fully-connected affinity.
Both cases are limited to adjust human-joint affinity for capturing the personalized poses of different actors in the video.
On the contrary,
our graph convolution with KNN can adaptively exploit most relevant joints to estimate a certain joint in 3D space,
which effectively reduces connection redundancy to boost performance.

\section{Method}
\label{Method}

In this section,
we first analyze spatial/temporal graph convolution with fixed human-joint affinity,
and explain how to design our dynamical human-joint affinity.
Then,
we integrate Dynamical Spatial/Temporal Graph convolution (DSG/DTG) to build up our Dynamical Graph Network (DG-Net) for 3D human pose estimation in videos.

\subsection{Dynamical Spatial Graph Convolution}
\label{Dynamical Spatial Graph Convolution}

For 3D pose estimation in videos,
spatial graph is to describe spatial relations between human joints at each frame.
Without loss of generality,
we denote it as $\mathcal{G}^{\mathcal{S}}=(\mathcal{V}^{\mathcal{S}}, \mathcal{E}^{\mathcal{S}})$,
where
$\mathcal{V}^{\mathcal{S}}$ and $\mathcal{E}^{\mathcal{S}}$ are respectively the node and edge sets.
Each node in $\mathcal{V}^{\mathcal{S}}$ refers to a human joint at a frame,
while
each edge in $\mathcal{E}^{\mathcal{S}}$ refers to the connection between two joints.

\textbf{Fixed Spatial Graph (FSG) Convolution}.
For frame $t$,
the node set $\mathcal{V}^{\mathcal{S}}$ corresponds to a feature matrix of human joints in this frame,
i.e.,
$\mathbf{X}_{t}=[\mathbf{x}_{t}^{1}, \mathbf{x}_{t}^{2},....,\mathbf{x}_{t}^{N}]\in \mathbb{R}^{N\times C_{x}}$,
where
$\mathbf{x}_{t}^{i}\in \mathbb{R}^{C_{x}}$ is the feature vector of the $i$-th human joint and $N$ is the number of joints.
Furthermore,
the edge set $\mathcal{E}^{\mathcal{S}}$ is represented by an affinity matrix $\mathbf{A}\in \mathbb{R}^{N\times N}$.
Traditionally,
the edge between joint $i$ and $j$ is defined by human skeleton.
Hence,
the affinity matrix $\mathbf{A}$ is fixed, i.e.,
\begin{equation}
\mathbf{A}(i,j)=\left\{
\begin{array}{ll}
1, &\mbox{$i$ and $j$ are connected in skeleton}\\
0, &\mbox{otherwise}
\end{array}
\right.
\label{SFA}
\end{equation}
Based on such fixed affinity,
spatial graph convolution performs message passing to update the node feature $\mathbf{X}_{t}$ as $\mathbf{Y}_{t}\in \mathbb{R}^{N\times C_{y}}$,
\begin{equation}
\mathbf{Y}_{t}=\sigma(\mathbf{A}\mathbf{X}_{t}\boldsymbol{\Theta}),
\label{FSGConv}
\end{equation}
where
$\boldsymbol{\Theta}\in \mathbb{R}^{C_{x}\times C_{y}}$ is the parameter matrix,
and
$\sigma$ is a nonlinear activation function,
e.g.,
ReLU.
Note that,
since $\mathbf{A}$ is based on the fixed human skeleton,
Eq.(\ref{FSGConv}) updates the feature of each human joint by integrating spatial context only from its physically-connected joints.
However,
a joint is not necessarily close to its physically-connected joints in different video frames,
e.g.,
for the \textit{SittingDown} case in the introduction,
\textit{wrist} is physically connected with \textit{elbow},
while its location is closer to and influenced more by \textit{hip} and \textit{pelvis}.
Only integrating context from \textit{elbow} may lead to unsatisfactory estimation of \textit{wrist}.

\textbf{Dynamical Spatial Graph (DSG) Convolution}.
To tackle the problem caused by the fixed affinity,
we propose to adaptively construct edges between joints for each frame.
Specifically,
for a human joint $i$ at frame $t$,
we find a set of its $K$ nearest joints $\Omega_{t}^{i}$,
according to the feature matrix of human joints $\mathbf{X}_{t}$ in this frame,
\begin{equation}
\Omega_{t}^{i}=\mbox{KNN}(\mathbf{x}_{t}^{i},\mathbf{X}_{t},K),
\label{KNN}
\end{equation}
where KNN refers to the $K$-Nearest-Neighbor algorithm,
and
we compute Euclidean distance between $\mathbf{x}_{t}^{i}$ and $\mathbf{X}_{t}$ to identify $\Omega_{t}^{i}$.
When joint $j$ belongs to $\Omega_{t}^{i}$,
there is an edge between joint $i$ and $j$.
Formally,
we can obtain a dynamical affinity matrix $\mathbf{D}_{t}$,
\begin{equation}
\mathbf{D}_{t}(i,j)=\left\{
\begin{array}{ll}
1,  & j \in \Omega_{t}^{i}\\
0,  &\mbox{otherwise}
\end{array}
\right.
\label{SDA}
\end{equation}
Moreover,
for joint $i$,
the importance of its neighbor joints in $\Omega_{t}^{i}$ can be different.
To take it into account,
we introduce a concise weighting mechanism to represent the importance score of joint $j\in \Omega_{t}^{i}$,
\begin{equation}
\mathbf{R}_{t}(i,j)=\gamma([\mathbf{x}_{t}^{i},~\mathbf{x}_{t}^{j}]),
\label{SWeight}
\end{equation}
where
$\gamma(\cdot)$ is a nonlinear mapping with the concatenated input of $[\mathbf{x}_{t}^{i}, \mathbf{x}_{t}^{j}]$.
In our experiment,
a fully-connected layer works well for $\gamma(\cdot)$.
Finally,
we obtain a weighted affinity via element-wise multiplication $\odot$,
\begin{equation}
\mathbf{B}_{t}=\mathbf{D}_{t}\odot\mathbf{R}_{t}.
\label{DSGA}
\end{equation}
This leads to our dynamical spatial graph convolution,
\begin{equation}
\mathbf{Y}_{t}=\sigma(\mathbf{B}_{t}\mathbf{X}_{t}\boldsymbol{\Phi}),
\label{DSGConv}
\end{equation}
where $\boldsymbol{\Phi}\in \mathbb{R}^{C_{x}\times C_{y}}$ is a parameter matrix.
Compared to the fixed and physical $\mathbf{A}$ in FSG (Eq. \ref{FSGConv}),
$\mathbf{B}_{t}$ in DSG is adaptively generated from human joint feature $\mathbf{X}_{t}$.
Hence,
it makes DSG robust to pose variations in different frames,
by dynamically finding and weighting the important spatial neighbors for each joint.




\subsection{Dynamical Temporal Graph Convolution}
\label{Dynamical Temporal Graph Convolution}

Spatial graph convolution mainly focuses on learning spatial relations between human joints in each frame,
while it ignores temporal relations between human joints in different frames.
To bridge this gap,
it is necessary to introduce temporal graph convolution.
Specifically,
we denote temporal graph as $\mathcal{G}^{\mathcal{T}}=(\mathcal{V}^{\mathcal{T}}, \mathcal{E}^{\mathcal{T}})$,
where
$\mathcal{V}^{\mathcal{T}}$ and $\mathcal{E}^{\mathcal{T}}$ are respectively the node and edge sets.
Without loss of generality,
we describe temporal graph at frame $t$ in the following.
First,
the node set $\mathcal{V}^{\mathcal{T}}$ consists of human joints in three adjacent frames,
$t-1$, $t$ and $t+1$.
This is mainly because we would like to leverage both \textit{forward} and \textit{backward} temporal context for frame $t$.
In this case,
the node set involves three matrices of human joint features,
i.e.,
$\mathbf{X}_{t-1}$, $\mathbf{X}_{t}$ and $\mathbf{X}_{t+1}$.
Second,
the edge set $\mathcal{E}^{\mathcal{T}}$ consists of human-joint connections between two adjacent frames. 
Hence,
it refers to two affinity matrices,
i.e.,
a \textit{forward} affinity matrix that describes connections between frame $t$ and $t+1$,
and
a \textit{backward} affinity matrix that describes connections between frame $t$ and $t-1$.

\textbf{Fixed Temporal Graph (FTG) Convolution}.
In the traditional temporal graph,
each human joint at frame $t$ is assumed to be connected with the same joint at frame $t+1$ and $t-1$.
As a result,
the forward and backward affinity matrices $\mathbf{P}_{t+1}$, $\mathbf{P}_{t-1} \in \mathbb{R}^{N\times N}$ are fixed and identical,
\begin{equation}
\mathbf{P}_{t+1}=\mathbf{P}_{t-1}=\mathbf{I},
\label{FMA}
\end{equation}
i.e.,
when joint $i$ at $t$ and joint $j$ at $t+1$ are corresponding to the same joint,
$\mathbf{P}_{t+1}(i,j)=1$.
Otherwise,
$\mathbf{P}_{t+1}(i,j)=0$.
It is the same case for $\mathbf{P}_{t-1}(i,j)$.
Based on such affinity matrices,
temporal graph convolution is actually reduced as the traditional temporal convolution,
\begin{equation}
\begin{aligned}
\mathbf{Y}_{t}
&=\sigma(\mathbf{X}_{t}\mathbf{W}_{x}+ \mathbf{P}_{t+1}\mathbf{X}_{t+1}\mathbf{W}_{f} + \mathbf{P}_{t-1}\mathbf{X}_{t-1}\mathbf{W}_{b})\\
&=\sigma(\mathbf{X}_{t}\mathbf{W}_{x}+ \mathbf{X}_{t+1}\mathbf{W}_{f} + \mathbf{X}_{t-1}\mathbf{W}_{b}),
\end{aligned}
\label{FTGconv}
\end{equation}
where
$\mathbf{W}_{x}, \mathbf{W}_{f}, \mathbf{W}_{b}\in \mathbb{R}^{C_{x}\times C_{y}}$ are the parameter matrices.
One can see that,
Eq. (\ref{FTGconv}) updates the feature of each human joint at $t$,
only by the features of the same joint at $t-1$ and $t+1$.
However,
such temporal context is often insufficient to reflect how this joint moves,
e.g.,
for \textit{Walking} in the introduction,
\textit{ankle} actually keeps the consistent movement with \textit{wrist} and \textit{shoulder} from $t$ to $t+1$.
If we integrate context only from \textit{ankle} at $t+1$,
temporal graph convolution tends to mistakenly estimate \textit{ankle} at $t$ to be forward.
\
\begin{figure*}[t]
\centering
\includegraphics[width=1\textwidth]{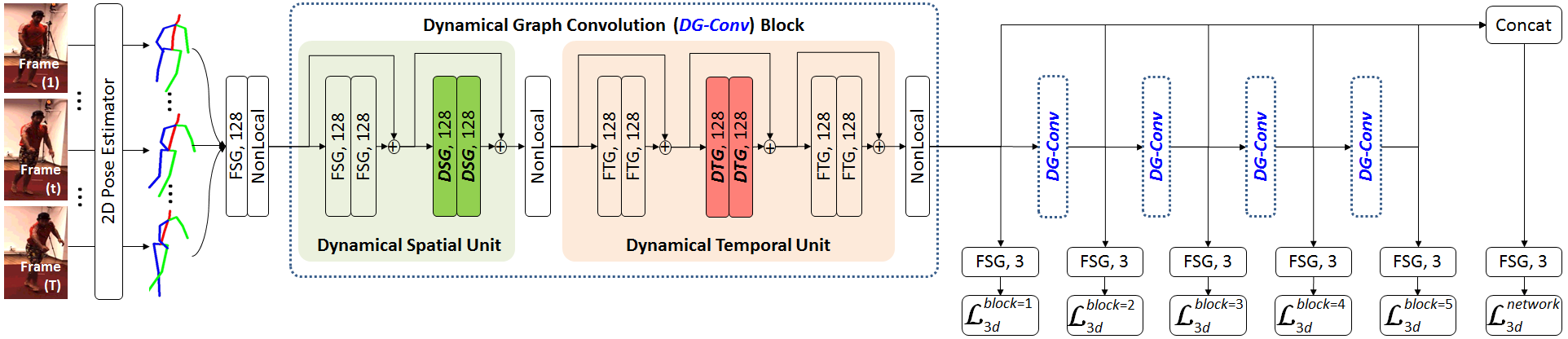}
\caption{Our DG-Net Architecture.
We design a dynamical graph convolution (DG-Conv) block to lift 2D pose to 3D pose.
Additionally,
we supervise DG-Net in a multi-level manner,
by leveraging the cooperation power of different DG-Conv blocks.
More details can be found in Section \ref{Dynamical Graph Network}.}
\label{Network-architecture}

\end{figure*}

\textbf{Dynamical Temporal Graph (DTG) Convolution}.
To deal with such problem in the temporal graph,
we propose to dynamically discover joint edges between two adjacent frames.
As mentioned in the introduction,
for a human joint at $t$,
other joints with similar movement (forward and/or backward) are the important temporal context to reduce motion uncertainty of this joint.
Hence,
we leverage the difference of joint features as guidance,
and find these contextual joints from frame $t+1$ and $t-1$.
Specifically,
for joint $i$ at frame $t$,
we use KNN to find a set of $K$ related joints at frame $t+1$ with similar forward motion,
\begin{equation}
\mho_{t+1}^{i}=\mbox{KNN}(\mathbf{x}_{t+1}^{i}-\mathbf{x}_{t}^{i},\mathbf{X}_{t+1}-\mathbf{X}_{t},K),
\label{KNND}
\end{equation}
where
we compute Euclidean distance of feature difference between this joint $\mathbf{x}_{t+1}^{i}-\mathbf{x}_{t}^{i}$ and all the joints in $\mathbf{X}_{t+1}-\mathbf{X}_{t}$,
in order to identify the forward set $\mho_{t+1}^{i}$.
Similarly,
we can find a set of $K$ joints at frame $t-1$ with similar backward motion,
$\mho_{t-1}^{i}=\mbox{KNN}(\mathbf{x}_{t}^{i}-\mathbf{x}_{t-1}^{i},\mathbf{X}_{t}-\mathbf{X}_{t-1},K)$.
Based on $\mho_{t+1}^{i}$ and $\mho_{t-1}^{i}$,
one can construct the associated affinity matrices that change over time,
e.g.,
when joint $j$ at frame $t+1$ belongs to the forward set $\mho_{t+1}^{i}$,
there is an edge between joint $i$ at frame $t$ and joint $j$ at frame $t+1$,
\begin{equation}
\mathbf{L}_{t+1}(i,j)=\left\{
\begin{array}{ll}
1,  & j\in \mho_{t+1}^{i}\\
0,  & \mbox{otherwise}
\end{array}
\right.
\label{FMAD}
\end{equation}
The backward case is similar,
i.e.,
$\mathbf{L}_{t-1}(i,j)=1$ when $j\in \mho_{t-1}^{i}$.
Otherwise,
$\mathbf{L}_{t-1}(i,j)=0$.
Moreover,
for joint $i$ at $t$,
the importance of joints in the forward set (or the backward set) can be different.
Hence,
like our dynamical spatial graph convolution before,
we introduce a weighting mechanism,
e.g.,
for joint $i$ at $t$,
we compute the importance score of joint $j$ in the forward set $\mho_{t+1}^{i}$,
\begin{equation}
\mathbf{H}_{t+1}(i,j)=\alpha([\mathbf{x}_{t}^{i},~\mathbf{x}_{t+1}^{j}]).
\label{weightDTG}
\end{equation}
Similarly,
we introduce a weighting mechanism for joint $j$ in the backward set $\mho_{t-1}^{i}$,
i.e.,
$\mathbf{H}_{t-1}(i,j)=\beta([\mathbf{x}_{t}^{i},~\mathbf{x}_{t-1}^{j}])$.
In our experiment,
a fully-connected layer works well for $\alpha(\cdot)$ and $\beta(\cdot)$ respectively.
Subsequently,
we obtain the weighted affinity matrices for both forward and backward cases,
\begin{equation}
\mathbf{Q}_{t+1}=\mathbf{L}_{t+1}\odot\mathbf{H}_{t+1},~~\mathbf{Q}_{t-1}=\mathbf{L}_{t-1}\odot\mathbf{H}_{t-1}.
\label{weightAD}
\end{equation}
This leads to our dynamical temporal graph convolution,
\begin{equation}
\mathbf{Y}_{t}
=\sigma(\mathbf{X}_{t}\mathbf{U}_{x}+ \mathbf{Q}_{t+1}\mathbf{X}_{t+1}\mathbf{U}_{f} + \mathbf{Q}_{t-1}\mathbf{X}_{t-1}\mathbf{U}_{b}),
\label{DTGconv}
\end{equation}
where
$\mathbf{U}_{x}, \mathbf{U}_{f}, \mathbf{U}_{b}\in \mathbb{R}^{C_{x}\times C_{y}}$ are the parameter matrices.
Compared to FTG in Eq. (\ref{FTGconv}),
our DTG contains two time-varying affinity matrices $\mathbf{Q}_{t+1}$ and $\mathbf{Q}_{t-1}$.
For each joint at $t$,
these matrices can effectively discover and weight its important temporal neighbors,
according to movement trends.
It allows our DTG to reduce motion uncertainty when estimating this joint.

\subsection{Dynamical Graph Network}
\label{Dynamical Graph Network}

In this section,
we integrate dynamical spatial and temporal graph convolution to build a Dynamical Graph Network (DG-Net) for 3D pose estimation in videos.
Specifically,
recent studies \cite{simple,semanticsgcn} have demonstrated that 3D pose can be lifted from 2D pose.
Hence,
we follow this concise style to design DG-Net.
As shown in Fig. \ref{Network-architecture},
we first use 2D pose estimator to predict 2D pose in each sampled frame,
and then design a dynamical graph convolution block to estimate 3D pose by learning spatio-temporal relations of human joints.

\textbf{Dynamical Graph Convolution (DG-Conv) Block}.
Since spatio-temporal factorization has been widely used for video learning \cite{tran2018closer,stgcn},
we build a DG-Conv block with two dynamical units.
First,
dynamical spatial unit learns spatial pose relations in each frame.
Then,
dynamical temporal unit learns temporal pose relations between different frames.
Moreover,
as shown in Fig. \ref{Network-architecture},
each dynamical unit is mixed with both fixed and dynamical graph convolution operations,
in order to take advantage of their complementary characteristics for pose feature enhancement.
For example,
dynamical spatial unit consists of FSG and DSG in Section \ref{Dynamical Spatial Graph Convolution}.
In this case,
this unit can first capture physical pose relations from human skeleton,
and then adjust relations dynamically according to pose variations in the video.
Dynamical temporal unit is a similar case.
Finally,
we apply the good practices of \cite{semanticsgcn} in our DG-Net,
e.g.,
we use FSG to map the input 2D pose of each frame into the 128-dim pose feature at the beginning.
In this case,
we can effectively exploit spatial and temporal neighbors for DSG/DTG,
by encoding joint location information in a flexible high-dimension space.
Additionally,
for each type of graph convolution in a dynamical unit,
we use the residual-style module built by two layers with 128 feature channels.
At the end of a DG-Conv block,
we map the final pose feature into the output 3D pose.
As suggested in \cite{semanticsgcn},
we also add a spatial non-local block after each dynamical unit to enhance holistic pose relations.
We repeat our DG-Conv block five times to increase network capacity.

\textbf{Multi-Level 3D Pose Supervision}.
We propose to supervise DG-Net in a multi-level manner.
First,
we use 3D pose supervision in the \textit{block level},
i.e.,
for each DG-Conv block,
we supervise the predicted 3D pose by ground truth,
\begin{equation}
\mathcal{L}_{3d}^{block} = \sum\nolimits_{t=1}^{T}\sum\nolimits_{i=1}^{N}(\hat{\boldsymbol{\rho}}^{block,i}_{t}-\boldsymbol{\rho}^{i}_{t})^{2},
\label{blocklevel}
\end{equation}
where
$\hat{\boldsymbol{\rho}}^{block,i}_{t}\in \mathbb{R}^{3}$ is the predicted 3D position of joint $i$ at frame $t$,
and
$\boldsymbol{\rho}^{i}_{t}\in \mathbb{R}^{3}$ is the corresponding ground truth.
In this case,
each block is regularized to be discriminative for predicting effective 3D pose.
Second,
we use 3D pose supervision in the \textit{network level},
i.e.,
we concatenate the output features of all DG-Conv blocks together.
Subsequently,
we map the concatenated feature into the predicted 3D pose,
and use ground truth to supervise it,
\begin{equation}
\mathcal{L}_{3d}^{network}  =\sum\nolimits_{t=1}^{T}\sum\nolimits_{i=1}^{N}(\hat{\boldsymbol{\rho}}^{network,i}_{t}-\boldsymbol{\rho}^{i}_{t})^2.
\label{networklevel}
\end{equation}
In this case,
DG-Net can fuse semantic representations of all DG-Conv blocks together to boost 3D pose estimation.
Finally,
we train our DG-Net with both losses,
\begin{equation}
\mathcal{L}_{3d}^{total} =\mathcal{L}_{3d}^{network}+ \lambda\sum\nolimits_{block=1}^{block=5}\mathcal{L}_{3d}^{block},
\label{alllevel}
\end{equation}
where
$\lambda$ is a weight coefficient.
This allows to train our DG-Net effectively,
by leveraging the cooperation power of both block and network levels of 3D pose supervision.
In the testing stage,
one can simply obtain the predicted 3D pose from the network-level output.

\begin{table*}[t]
\begin{center}
\resizebox{\textwidth}{!}{
\renewcommand\tabcolsep{3pt}
\begin{tabular}{l| c | c |c |c |c |c |c |c |c |c |c |c |c |c |c |c |c}
\hline\hline
Protocol$1$ (mm)
&$T$  &  Direct.       & Discuss    &  Eat       & Greet            &  Phone             &  Photo       &  Pose        &  Purch. &  Sitting &  SitD.  &  Smoke     &  Wait      &  WalkD   &  Walk.   &WalkT.    &  Avg\\
\hline\hline

Martinez et al.\cite{simple}
& 1 &51.8     &56.2     &58.  &59.0      &69.5          &78.4     &55.2     &58.1&74.0 &94.6&62.3   &59.1  &65.1 &49.5 &52.4 &62.9\\
Fang et al.\cite{rnnpose}
& 1 &50.1       &54.3     &57.0   &57.1     &66.6 &73.3     &53.4   &55.7&72.8 &88.6&60.3   &57.7   &62.7 &47.5 &50.6 &60.4\\
Yang et al.\cite{adversialpose}
& 1 &51.5    &58.9   &50.4  &57.0     &62.1          &65.4    &49.8    &52.7 &69.2 &85.2 &57.4 &58.4  &43.6 &60.1 &47.7 &58.6\\
Zhao et al.\cite{semanticsgcn}
& 1 &47.3    &60.7     &51.4  &60.5    &61.1          &49.9   &47.3    &68.1 &86.2 &55.0 &67.8 &61.0 &\textbf{42.1} &60.6&45.3&57.6\\
Pavlakos et al. \cite{ordinal}
& 1 &48.5      &54.4     &54.4  &52.0      &59.4          &65.3     &49.9     &52.9&65.8 &71.1&56.6   &52.9   &60.9 &44.7 &47.8 &56.2\\
Ci et al.\cite{Ci_2019_ICCV}
& 1 &46.8 &52.3     &44.7  &50.4     &52.9           &68.9     &49.6     &46.4&60.2&78.9&51.2   &50.0   &54.8 &40.4 &43.3 &52.7\\
Wang et al.\cite{Wang_2019_ICCV}
& 1 &44.7&48.9    &47.0  &49.0      &56.4           &67.7   &48.7     &47.0&63.0 &78.1&51.1   &50.1   &54.5 &40.1 &43.0 &52.6\\
Lee et al.\cite{Lee_2018_ECCV}
&  3  &40.2  &49.2 &47.8  &52.6      &50.1          &75.0   &50.2    &43.0&55.8 &73.9&54.1   &55.6   &43.3 &58.2 &43.3 &52.8\\
 Marcard et al.\cite{eccv2018temporal}
&  5  &44.2      &46.7 &52.3   &49.3     &59.9          &59.4    &47.5    &46.2&59.9 &65.6&55.8   &50.4   &52.3 &43.5 &45.1 &51.9\\
Cai et al.\cite{Cai_2019_ICCV}
&  7  &44.6      &47.4      &45.6   &48.8      &50.8          &59.0     &47.2    &43.9&57.9 &61.9&49.7  &46.6  &51.3 &37.1 &39.4 &48.8\\
Xu et al.\cite{Xu_2020_CVPR}
&7 &\textbf{38.2} &\textbf{44.4} &42.8&\textbf{43.7} &47.6 &60.3 &\textbf{42.0} &45.4 &\textbf{53.2} &60.8 &\textbf{46.4} &43.5 &48.5 &34.6 &38.6 &46.3 \\
Liu et al.\cite{liu2020gast} 
&27 &44.9&46.7 &41.9 &45.6 &47.9 &56.1 &44.2 &45.5 &57.1 &59.1 &46.8 &43.5 &47.5 &\textbf{32.6} &\textbf{33.1} &46.2\\
Dario et al.\cite{pavllo:videopose3d:2019}
& 243 &45.2      &46.7 &43.3  &45.6 &48.1 &55.1&44.6&44.3&57.3 &65.8&47.1&44.0&49.0&32.8 &33.9 &46.8\\
\hline
Our DG-Net			
& 4   &41.5    &46.6     &\textbf{41.0}&44.3&\textbf{47.1}&\textbf{54.1} &44.2&\textbf{42.5}&54.9 &\textbf{58.8}&46.9&\textbf{43.1}&46.9&\textbf{32.6} &35.6&\textbf{45.3}\\
\hline\hline

Ci et al.\cite{Ci_2019_ICCV}$\dagger$
& 1 &36.3    &38.8   &29.7    &37.8    &34.6    &42.5   &39.8  &32.5   &\textbf{36.2}   &\textbf{39.5}   &34.4  &38.4   &38.2  &31.3   &34.2   &36.3\\
Cai et al.\cite{Cai_2019_ICCV}$\dagger$
& 3  &32.9   &38.7  &32.9  &37.0   &37.3  &44.8  &38.7  &36.1   &41.0  &45.6   &36.8 &37.7  &37.7  &29.5  &31.6 &37.2\\
 Marcard et al.\cite{eccv2018temporal}$\dagger$
& 5   &35.2   &40.8 &37.2  &37.4   &43.2   &44.0  &38.9 &35.6   &42.3  &44.6   &39.7   &39.7 &40.2 &32.8 &35.5  &39.2\\
Pavllo et al.\cite{pavllo:videopose3d:2019}$\dagger$
& 243 &----- &----- &----- &-----  &----- &-----  &----- &----- &----- &-----  &----- &----- &----- &----- &----- &37.1\\
Liu et\cite{Liu_2020_CVPR}$\dagger$
&243 &32.3 &35.2&33.3&35.8 &35.9 &41.5 &33.2 &32.7 &44.6 &50.9 &37.0 &32.4 &37.0 &25.2 &27.2 &34.7\\
\hline
Our DG-Net$\dagger$
&4    &\textbf{28.5}   &\textbf{33.5}   &\textbf{28.1}   &\textbf{28.9}    &\textbf{32.6}   &\textbf{35.5}  &\textbf{33.3}  &\textbf{30.0}    &37.4  &39.9  &\textbf{31.4}   &\textbf{30.2}  &\textbf{29.5}   &\textbf{23.9} &\textbf{25.5} &\textbf{31.2}\\
\hline\hline			
\end{tabular}
}
\end{center}
\caption{SOTA Comparison on Human3.6M under Protocol 1 (mm).
The mark $ \dagger$ refers to using GT 2D pose as input of methods.
$T$ is the input sequence length.
Note that,
our 4-frame DG-Net outperforms 243-frame SOTA method \cite{pavllo:videopose3d:2019}.}
\label{protocol1}
\end{table*}

\begin{table*}[t]
\begin{center}
\resizebox{\textwidth}{!}{
\renewcommand\tabcolsep{3pt}
\begin{tabular}{l| c | c |c |c |c |c |c |c |c |c |c |c |c |c |c |c |c}
\hline\hline
Protocol$2$ (mm)
&$T$  &  Direct.       & Discuss    &  Eat       & Greet            &  Phone             &  Photo       &  Pose        &  Purch. &  Sitting &  SitD.  &  Smoke     &  Wait      &  WalkD   &  Walk.   &WalkT.    &  Avg\\
\hline\hline
Pavlakos et al.\cite{volumetric}
& 1 &-----     &----- &----- &-----       &-----         &-----   &-----  &----- &----- &----- &----- &----- &----- &----- &----- &51.9\\
Martinez et al.\cite{simple}
& 1 &39.5       &43.2      &46.4   &47.0      &51.0           &56.0     &41.4     &40.6&56.5 &59.4&49.2   &45.0   &49.5 &38.0 &43.1 &47.7\\
Fang et al.\cite{rnnpose}
& 1 &38.2       &41.7      &43.7   &44.9      &48.5           &55.3     &40.2     &38.2&54.5 &64.4&47.2   &44.3   &47.3 &36.7 &41.7 &45.7\\

Pavlakos et al. \cite{ordinal}
& 1 &34.7       &39.8      &41.8   &38.6      &42.5           &47.5     &38.0     &36.6 &50.7 &56.8&42.6   &39.6   &43.9 &32.1 &36.5 &41.8\\
Ci et al.\cite{Ci_2019_ICCV}
& 1 &36.9       &41.6      &38.0   &41.0      &41.9           &51.1     &38.2     &37.6&49.1 &62.1&43.1   &39.9   &43.5 &32.2 &37.0 &42.2\\
Wang et al.\cite{Wang_2019_ICCV}
& 1 &32.8&36.8      &42.5   &38.5      &42.4           &49.0     &35.4     &34.3&53.6 &66.2&46.5   &34.1   &42.3 &30.0 &39.7 &42.2\\
Lee et al.\cite{Lee_2018_ECCV}
&  3  &34.9  &\textbf{35.2} &43.2   &42.6      &46.2           &55.0     &37.6     &38.8&50.9 &67.3&48.9   &35.2   &\textbf{31.0} &50.7 &34.6 &43.4\\
Marcard et al.\cite{eccv2018temporal}
&  5  &36.9       &37.9 &42.8   &40.3      &46.8           &46.7     &37.7     &36.5&48.9 &52.6&45.6   &39.6   &43.5 &35.2 &38.5 &42.0\\
Cai et al.\cite{Cai_2019_ICCV}
&  7  &35.7       &37.8      &36.9   &40.7      &39.6           &45.2     &37.4     &34.5&46.9 &50.1&40.5   &36.1   &41.0 &29.6 &33.2 &39.0\\
Xu et al.\cite{Xu_2020_CVPR}
&7 &\textbf{31.7} &35.3 &35.0 &35.3 &36.9 &44.2 &32.0 &33.8 &\textbf{42.5} &49.3 &37.6 &33.4 &39.6 &27.6 &32.5 &36.7 \\
Liu et al.\cite{liu2020gast} 
&27 &34.0 &36.6 &33.5 &37.2 &36.4 &42.7 &34.1 &34.4 &45.5 &47.0 &\textbf{37.2} &33.0 &36.6 &\textbf{24.9} &\textbf{26.9} &36.0\\
Pavllo et al.\cite{pavllo:videopose3d:2019}
& 243 &34.1       &36.1 &34.4   &37.2 &36.4 &42.2&34.4&33.6&45.0 &52.5&37.4&33.8&37.8&25.6 &27.3 &36.5\\
\hline
Our DG-Net			
& 4   &32.9      &36.1      &\textbf{32.8}&\textbf{35.1}&35.8&42.1 &\textbf{31.9}&\textbf{32.2}&43.9 &\textbf{46.4}&38.0&\textbf{32.9}&35.7 &25.5 &29.6&\textbf{35.4}\\
\hline\hline
Martinez et al.\cite{simple} $\dagger$
& 1 &-----  &----- &-----  &-----  &-----  &-----  &----- &-----  &----- &-----  &-----  &----- &----- &----- &----- & 37.1\\
Ci et al.\cite{Ci_2019_ICCV}$\dagger$
& 1 &24.6    &28.6   &24.0    &27.9    &27.1    &31.0    &28.0   &25.0    &31.2   &35.1    &27.6    &28.0   &29.1   &24.3   &26.9   &27.9\\

\hline
Our DG-Net$\dagger$
&4    &\textbf{20.5}    &\textbf{26.0}   &\textbf{21.8}    &\textbf{23.0}    &\textbf{23.7}    &\textbf{26.7}    &\textbf{25.2}   &\textbf{23.1}    &\textbf{29.5}   &\textbf{32.5}    &\textbf{24.2}    &\textbf{23.3}  &\textbf{23.7}    &\textbf{18.7}   &\textbf{19.6} &\textbf{24.1}\\
\hline\hline			
\end{tabular}
}
\end{center}
\caption{SOTA Comparison on Human3.6M under Protocol2 (mm).
The mark $ \dagger$ refers to using GT 2D pose as input of methods.
$T$ is the input sequence length.
Note that,
our 4-frame DG-Net outperforms 243-frame SOTA method \cite{pavllo:videopose3d:2019}.}
\label{protocol2}
\vspace{-0.3cm}	
\end{table*}

\begin{table}[t]
\begin{center}
\resizebox{0.48\textwidth}{!}{	
\renewcommand\tabcolsep{3pt}	
\begin{tabular}{l|c|c c c |c c c |c }
\hline\hline
\multirow{2}{*}{Protocol$2$ (mm)} & \multirow{2}{*}{$T$} & \multicolumn{3}{c|}{Walking} &\multicolumn{3}{c|}{Joging} &\multirow{2}{*}{Avg}  \\
                  &     & S1 &S2 &S3                   &S1 &S2 &S3                  & \\
\hline\hline
Yasin et al.\cite{Yasin_2016_CVPR}
&1 &35.8 &32.4 &41.6 &46.6 &41.4 &35.4 &38.9\\
Lin et al.\cite{Lin_2017_CVPR}
&1&26.5 &20.7 &38.0 &41.0 &29.7 &29.1 &30.8\\
Moreno-Noguer et al.\cite{distance}
&1 &19.7 &13.0 &\textbf{24.9} &39.7 &20.0 &21.0 &26.9\\
Pavlakos et al.\cite{volumetric}
&1 &22.1 &21.9 &29.0 &29.8 &23.6 &26.0 &25.5\\
Martinez et al.\cite{simple}
&1 & 19.7 &17.4 &46.8 &26.9 &18.2 &18.6 &24.6\\
Fang et al.\cite{rnnpose}
&1 &19.4 &16.8 &37.4 &30.4 &17.6 &16.3 &22.9\\
Marcard et al.\cite {eccv2018temporal}
&5   &19.1 &13.6 &43.9 &23.2 &16.9 &15.5 &22.0\\
Pallvo et al.\cite{pavllo:videopose3d:2019}
&27  &14.5 &10.5 &47.3 &21.9 &13.4 &13.9 &20.2\\
\hline
Our DG-Net
&4   &\textbf{13.7} &\textbf{9.5} &47.1 &\textbf{21.0} &\textbf{12.6} &\textbf{13.4} &\textbf{19.5}\\			
\hline\hline
\end{tabular}
}		
\end{center}
\caption{SOTA Comparison on HumanEva-I under Protocol2 (mm).}
\label{humaneva}
\vspace{-0.4cm}
\end{table}

\begin{table}[t]
\begin{center}
\resizebox{0.48\textwidth}{!}{		
\begin{tabular}{l |c | c c c}
\hline\hline
2D-to-3D Pose Lifters & $T$   & PCK$\uparrow$   & AUC$\uparrow$  & MPJPE (mm)$\downarrow$   \\
\hline\hline
Mono et al.\cite{mono_3dhp2017}
&1 &75.7 &39.3 &-\\
Mehta et al.\cite{VNect_SIGGRAPH2017}
& 1&   79.4 &41.6     &-\\
Lin et al.\cite{lin2019trajectory}
&25&83.6    &  51.4   &79.8\\
Lin et al.\cite{lin2019trajectory}
&50 &   82.4 &49.6     &81.9\\ \hline
Our DG-Net
& 4&\textbf{87.5}    &\textbf{53.8}   &\textbf{76.0}\\			
\hline\hline
\end{tabular}
}		
\end{center}
\caption{SOTA Comparison on MPI-INF-3DHP under PCK, AUC and MPJPE (mm).}
\label{MPI}
\vspace{-0.4cm}
\end{table}

\section{Experiments}

\textbf{Datasets}.
We perform DG-Net on three widely-used benchmarks in 3D pose estimation,
including
Human3.6M \cite{h36m_pami},
HumanEva-I \cite{humaneva},
and
MPI-INF-3DHP \cite{mono_3dhp2017}.
Human3.6M consists of 3.6 million images from 4 different cameras where 7 actors perform 15 activities in the indoor environment.
Following the previous works \cite{pavllo:videopose3d:2019,Ci_2019_ICCV,Cai_2019_ICCV},
we use subjects 1, 5, 6, 7, 8 for training and subjects 9, 11 for testing.
The evaluation metric is the mean per joint position error (MPJPE) between ground truth and estimated 3D pose,
which refers to Protocol1.
We also report the results of Protocol2,
where
estimated 3D pose is aligned to ground truth via a rigid transformation.
Additionally,
following \cite{pavllo:videopose3d:2019},
we extract the input 2D pose of Human3.6M from cascaded pyramid network \cite{cpn}.
HumanEva-I contains video sequences of 3 subjects with 6 actions which are recorded by 3 cameras.
Following \cite{pavllo:videopose3d:2019,simple}, 
we train DG-Net and test Protocol2 results for each action.
Moreover,
as suggested in \cite{pavllo:videopose3d:2019},
we extract the input 2D pose of HumanEva-I from Mask R-CNN \cite{maskrcnn}. 
MPI-INF-3DHP is captured with a multi-view setup,
where
subjects do not wear visible markers.
Following \cite{VNect_SIGGRAPH2017},
the evaluation metrics of MPI-INF-3DHP are
MPJPE,
PCK (percentage of correct keypoints),
and
AUC (area under curve).

\textbf{Implementation Details}.
Unless stated otherwise,
we implement DG-Net as follows.
The number of neighbors $K$ in DSG and DSG are respectively 3 and 4.
The weighting function in Eq. (\ref{SWeight}) and (\ref{weightDTG}) refer to a fully-connected layer.
DG-Net contains 5 DG-Conv blocks.
The number of input frames is 4.
In the loss function of Eq. (\ref{alllevel}),
$\lambda$ is set as $0.1$.
We implement DG-Net in PyTorch,
where
we use Adam optimizer with an initial learning rate of 0.001 and the exponential decay. 
We train our model for 200 epochs using mini-batches of size 64 on a single GTX 1080 GPU.

\begin{figure}[t]
\centering
\includegraphics[width=0.47\textwidth]{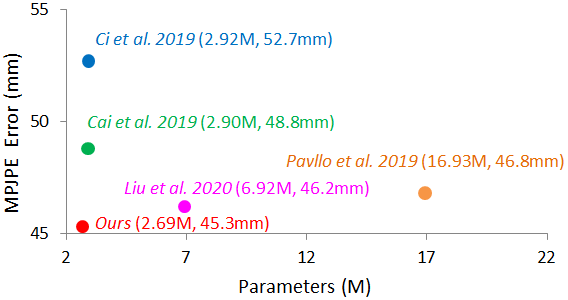}
\caption{MPJPE vs. Model Size (Human3.6M).
Comparing with Pavllo et al. \cite{pavllo:videopose3d:2019},
our DG-Net achieves a smaller MPJPE but with $84.1\%$ parameter reduction.}
\label{TOFFs}
\end{figure}

\subsection{SOTA Comparison}

\textbf{Human3.6M}.
We compare our approach to the state-of-the-arts on the Human3.6M dataset.
The results of protocol 1 and 2 are shown in Table~\ref{protocol1} and \ref{protocol2}.
One can see that,
our DG-Net outperforms the SOTA method \cite{pavllo:videopose3d:2019},
with only $4/243=1.6\%$ input frames.
Moreover,
when using ground truth 2D pose,
our DG-Net achieves $3.5$ mm (protocol 1) and $3.8$ mm (protocol 2) estimation improvement,
compared to the SOTA approaches \cite{Ci_2019_ICCV,Liu_2020_CVPR}.
Finally,
our DG-Net obtains the smaller estimation error on complex actions (e.g., Sit Down, Photo) that contain high depth ambiguity and motion uncertainty of human joints.
It indicates that,
our method adaptively learns spatio-temporal joint relation to capture human pose in different actions.

\textbf{HumanEva-I}.
Following the previous methods \cite{Yasin_2016_CVPR,Lin_2017_CVPR,distance,volumetric,simple,rnnpose,eccv2018temporal,pavllo:videopose3d:2019},
we report results on Walking and Jogging in three subjects of HumanEva-I by using protocol 2.
As shown in Table~\ref{humaneva},
we achieve a better overall performance.
Note that,
the high error on S3 (Walk) of HumanEva-I is due to corrupted mocap data,
as indicated by \cite{pavllo:videopose3d:2019}.

\textbf{MPI-INF-3DHP}.
As suggested in \cite{lin2019trajectory},
we mainly compare the previous 2D-to-3D pose lifters for MPI-INF-3DHP,
in order to show generalization capacity of our model.
In this case,
the ground truth 2D pose is given as input for all the methods.
Clearly, as shown in Table \ref{MPI},
our DG-Net outperforms other lifters on all the evaluation metrics.

\textbf{Model Size and Inference Time}.
As suggested in \cite{Ci_2019_ICCV,Liu_2020_CVPR},
we show tradeoff between model size and estimation error in Fig. \ref{TOFFs}.
It is worth mentioning that,
we use multi-level 3D pose supervision in the training phase,
while
we simply obtain the predicted 3D pose from the network-level output in the testing phase.
Hence,
in the testing phase,
we do not need the final FSG module (that is just above each block-level loss in Fig. \ref{Network-architecture}) to generate the block-level output.
As a result,
the size of model parameters (2.39M) in the testing phase is smaller than that (2.69M) in the training phase.
However,
to keep consistency with other approaches,
we still report the model size of training phase in Fig. \ref{TOFFs},
where
our DG-Net achieves the best balance among the related approaches.
Additionally,
we further show the inference time,
by comparing our method with the SOTA approach \cite{pavllo:videopose3d:2019}.
As shown in Table \ref{time},
the inference time of\cite{pavllo:videopose3d:2019} is 2.3 ms/frame,
while that of our method is 0.5 ms/frame,
by using the same GTX 1080 GPU.
It proves that our method is not only effective but also efficient.
Moreover,
it is worth mentioning that,
KNN would bring little computation cost in our model,
and $K$ is not to be large to achieve good performance (e.g., $K=4$ for each joint in our experiment).
To validate it,
we further show the inference time of our method without KNN.
One can see that,
DG-Net without KNN spends 0.41ms/frame on inference with MPJPE 52.1 mm,
while
the proposed DG-Net spends 0.52ms/frame on inference with MPJPE 45.5 mm.
It clearly demonstrates that,
KNN in our design brings little more computation but leads to much better estimation.


\begin{table}[t]
\begin{center}
\begin{tabular}{l|cc}
\hline\hline
Human3.6M                                       &Inference Time &  MPJPE      \\
\hline
Pallvo et al. \cite{pavllo:videopose3d:2019}    &2.33ms/frame    & 46.8 mm\\
\hline
DG-Net without KNN                              &0.41ms/frame    &52.1 mm\\
\hline
Our DG-Net                                      &\textbf{0.51}ms/frame    & \textbf{45.3} mm\\
\hline\hline
\end{tabular}
\caption{Inference Time.}
\label{time}
\end{center}
\vspace{0.1cm}
%
%
%
%
%
\begin{center}
\begin{tabular}{l|cccc}
\hline\hline
$K$ in DSG &1      &3      &5      &7           \\
\hline
Our DG-Net     & 47.3 mm & 45.3 mm & 48.2 mm & 50.8 mm \\
\hline\hline
$K$ in DTG &1      &3      &5      &7          \\
\hline
Our DG-Net     & 48.0 mm & 46.5 mm  & 47.3 mm & 49.2 mm \\
\hline\hline
\end{tabular}
\caption{No. of joint neighbors $K$ used in DSG and DTG.}
\label{M_FF}
\end{center}
\vspace{0.1cm}
%
%
%
%
%
%
\begin{center}
\begin{tabular}{l| c c c }
\hline\hline
\diagbox{Graph}{Weighting}   & W/O    & FC  & EG    \\
\hline\hline
Random  &83.4 mm &80.5 mm &80.9 mm           \\
Fixed   &52.7 mm &50.3 mm &51.4 mm           \\
Full    &58.8 mm &56.3 mm &57.3 mm \\
Symmetry&53.2 mm &53.0 mm &53.0 mm\\
Precomputed &50.4 mm &48.1 mm &48.8 mm\\
\hline
Our Dynamical &48.1 mm &45.3 mm &46.9 mm\\
\hline\hline
\end{tabular}
\caption{Graph Connection Styles \& Weighting Functions in DSG and DTG.
W/O: Without Weighting.
FC: Fully-Connected Layer.
EG: Embedded Gaussian Kernel.}
\label{hhhhhh}
\end{center}
%
\vspace{0.1cm}
\begin{center}
\begin{tabular}{l|c}
\hline\hline
Dynamical Spatial Unit                                                       & MPJPE (mm)    \\
\hline\hline
\textit{FSG}                                                                         &   57.9       \\
\textbf{\textit{FSG+DSG}}                                                    &  \textbf{55.8}        \\
\textit{FSG+FSG}                                                                  &   57.4      \\
\textit{FSG+DSG+DSG}                                                                  &   57.4       \\
\textit{FSG+DSG+FSG}                                                                 &    56.5      \\
\hline\hline
Dynamical Temporal Unit                                                      & MPJPE (mm)   \\
\hline\hline
\textbf{\textit{FSG+DSG}}+\texttt{FTG}                                                            &  52.0        \\
\textbf{\textit{FSG+DSG}}+\texttt{FTG+DTG}                                                        &     47.0     \\
\textbf{\textit{FSG+DSG}}+\texttt{FTG+FTG}                                                        &     51.3   \\
\textbf{\textit{FSG+DSG}}+\texttt{FTG+DTG+DTG}                                                    &    47.6      \\
\textbf{\textit{FSG+DSG}}+\textbf{\texttt{FTG+DTG+FTG}}                                           &  \bf{45.3}        \\
\textbf{\textit{FSG+DSG}}+\texttt{FTG+DTG+FTG+DTG}                                                &  46.1      \\
\textbf{\textit{FSG+DSG}}+\texttt{FTG+DTG+FTG+FTG}                                                &  45.5       \\
\hline\hline
\end{tabular}%
\caption{DG-Conv Block.}
\label{tab:dgcov}
\end{center}
\vspace{-0.7cm}
\end{table}

\begin{figure*}[t]
\centering
\includegraphics[width=1\textwidth]{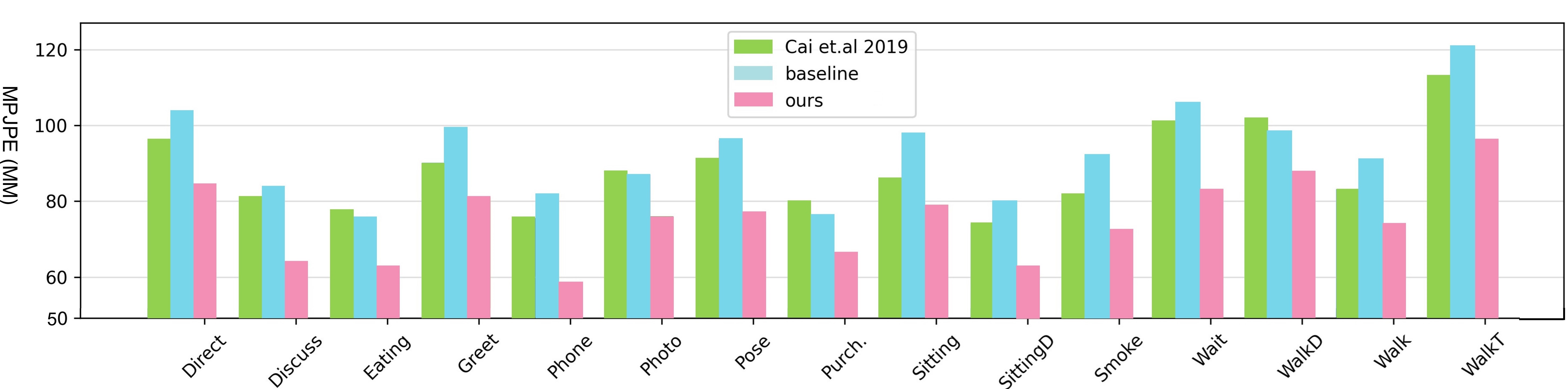}
\caption{Cross Actions Generalization Ability.
We train
baseline,
2D-to-3D pose lifter \cite{Cai_2019_ICCV},
and our DG-Net on each of the 15 actions in Human3.6M.
Then,
we test these models on all the actions.
X-axis indicates the action used in training.
Our method achieves smaller MPJPE results in all actions}
\label{VC333}
\end{figure*}

 \begin{table}[t]
 \begin{center}
 \begin{tabular}{c|c}
 \hline
 DG-Conv Blocks & MPJPE (mm)  \\
 \hline
 1             &   60.7  \\
 3             &   47.0  \\
 5             &    \bf{45.3} \\
 \hline
 Input Frames ($T$) & MPJPE (mm)  \\
 \hline
 2             &  51.2  \\
 3             &  48.1  \\
 4             &  \textbf{45.3}  \\
 5             &  \textbf{45.3}   \\
 \hline
 \end{tabular}%
\caption{No. of DG-Conv Blocks \& Input Frames.}
\label{tab:blockframe}
\end{center}
\vspace{0.3cm}
%
%
%
\begin{center}
			\begin{tabular}{l|c c c |c  }
			  \hline\hline
			 Train/Test (Protocol2 (mm)) &\cite{distance} & \cite{simple} & \cite{eccv2018temporal} &Our\\
			  \hline\hline
			  GT / GT   &62.17 &37.10 &31.6 &\textbf{24.17}\\
			  GT / GT+$\mathcal{N}(0,5)$ & 67.11 &46.65 &37.46 &\textbf{31.86}\\
			   GT / GT+$\mathcal{N}(0,10)$ &79.12 &52.84 &49.41 &\textbf{38.63}\\
			    GT / GT+$\mathcal{N}(0,15)$ & 96.08 &59.97 &61.80 &\textbf{46.92}\\
		     GT / GT+$\mathcal{N}(0,20)$ & 115.55 &70.24 &73.65 &\textbf{56.33}\\
			 \hline\hline
			\end{tabular}
\end{center}
\caption{Robustness-to-Noise.}
\label{noise}
\vspace{-0.3cm}
\end{table}

\subsection{Ablation Study}

\textbf{Number of joint neighbors $K$ in DSG and DTG}.
Without loss of generality,
we investigate ablation studies on Human3.6M.
When we change $K$ in DSG (or DTG),
we fix $K$ in DTG (or DSG) as 4 (or 3).
As expected,
when we increase $K$ in DSG (or DTG) in Table \ref{M_FF},
the estimation error of DG-Net first decreases and then increases.
The main reason is that,
when a joint is connected with too few (or many) neighbors,
its pose context tends to be insufficient (or noisy).
Hence,
we choose the moderate $K$ (3/4 for DSG/DTG) in our experiments.
As the layer is going deeper,
the receptive field of spatial/temporal affinity is getting bigger,
which can boost pose estimation by learning discriminative clues from local to global context of human joints.


\textbf{Graph Connection Styles \& Weighting Functions}.
For graph connection style in DSG and DTG,
we investigate six settings.
The random setting is to randomly select $K$ neighbors for each joint.
The full setting is to connect each joint with all other joints.
The fixed setting is to connect joints according to human skeleton.
The symmetry setting refers to the symmetrical connection of joints in \cite{rnnpose}.
The precomputed setting is to find $K$ neighbors for each joint by the estimated 2D pose,
and then fix these connections in DSG and DTG.
The dynamical setting is our design.
For the weighting function,
we investigate three settings.
Except the without setting,
we use a FC layer or the kernel in \cite{nonlocal} to obtain the importance score.
We denote them as the fully connected or embedded gaussian settings.
As shown in Table \ref{hhhhhh},
the dynamical setting achieves the best among all the graph styles,
showing the effectiveness of our design.
Moreover,
the fully connected setting of weighting mechanism is preferable.
Hence,
we choose this setting in our experiment.
Additionally,
the full setting of graph connection with various weighting functions is actually the case of graph convolution with self-attention.
As shown in Table\ref{M_FF},
our dynamical setting with KNN clearly outperforms this setting.
The main reason is that graph convolution with soft affinity matrix learned by self-attention is often redundant and noisy,
since all the joints are leveraged to estimate a certain joint.
Alternatively,
KNN can adaptively exploit most relevant joints to estimate a certain joint in 3D space,
which effectively reduces connection redundancy to boost performance.

\textbf{
Dynamical Spatial/Temporal Units.}
We gradually introduce our DSG and DTG operations in DG-Conv Block.
As shown in Table \ref{tab:dgcov},
the performance achieves the best with
(dynamical spatial unit: FSG+DSG)+(dynamical temporal unit: FTG+DTG+FTG).
Hence,
we choose this setting in our experiment.
 Note that,   FSG+DSG+DSG and FSG+DSG+FSG settings are used to illustrate that FSG+DSG is empirically the best. Adding more spatial graph convolution (either fixed or dynamical) on FSG+DSG would increase unnecessary model complexity to reduce performance. Hence, both cases are the poor settings. FSG+DSG+DSG is relatively worse than FSG+DSG+FSG, since it may introduce more complexity due to extra dynamical spatial affinity.

\begin{table}[t]
\begin{center}
\begin{tabular}{l|c|c}
\hline\hline
Supervision            & Network Loss   & Total Loss     \\
\hline
Our DG-Net             &    48.7  mm      &    \textbf{45.3}  mm     \\
\hline\hline
Nonlocal               & Without   & With     \\
\hline
Our DG-Net             &  50.1  mm        & \textbf{45.3}   mm       \\
\hline\hline
Coordinate Operation   & Non-Balanced   & Balanced    \\
\hline
Our DG-Net             &   46.1   mm      &  \textbf{45.3}   mm      \\
\hline\hline
Choice of 2D Estimator &SH \cite{alej2016stacked} &  CPN \cite{cpn}    \\
\hline
Pavllo et al.\cite{pavllo:videopose3d:2019} &53.4  mm &46.8 mm\\
Our DG-Net             &   \textbf{49.9} mm     &\textbf{45.3}  mm         \\
\hline\hline
DG Form   & Unified   & Factorized    \\
\hline
Our DG-Net             &   48.9   mm      &  \textbf{45.3}   mm      \\
\hline\hline
\end{tabular}%
\caption{Detailed Designs (Supervision, Nonlocal, Coordinate Operation, Choice of 2D Pose Estimator, DG Form).}
\label{tab:detail}
\end{center}
\vspace{-0.5cm}
\end{table}

\begin{figure*}[t]
\centering
\includegraphics[width=0.98\textwidth]{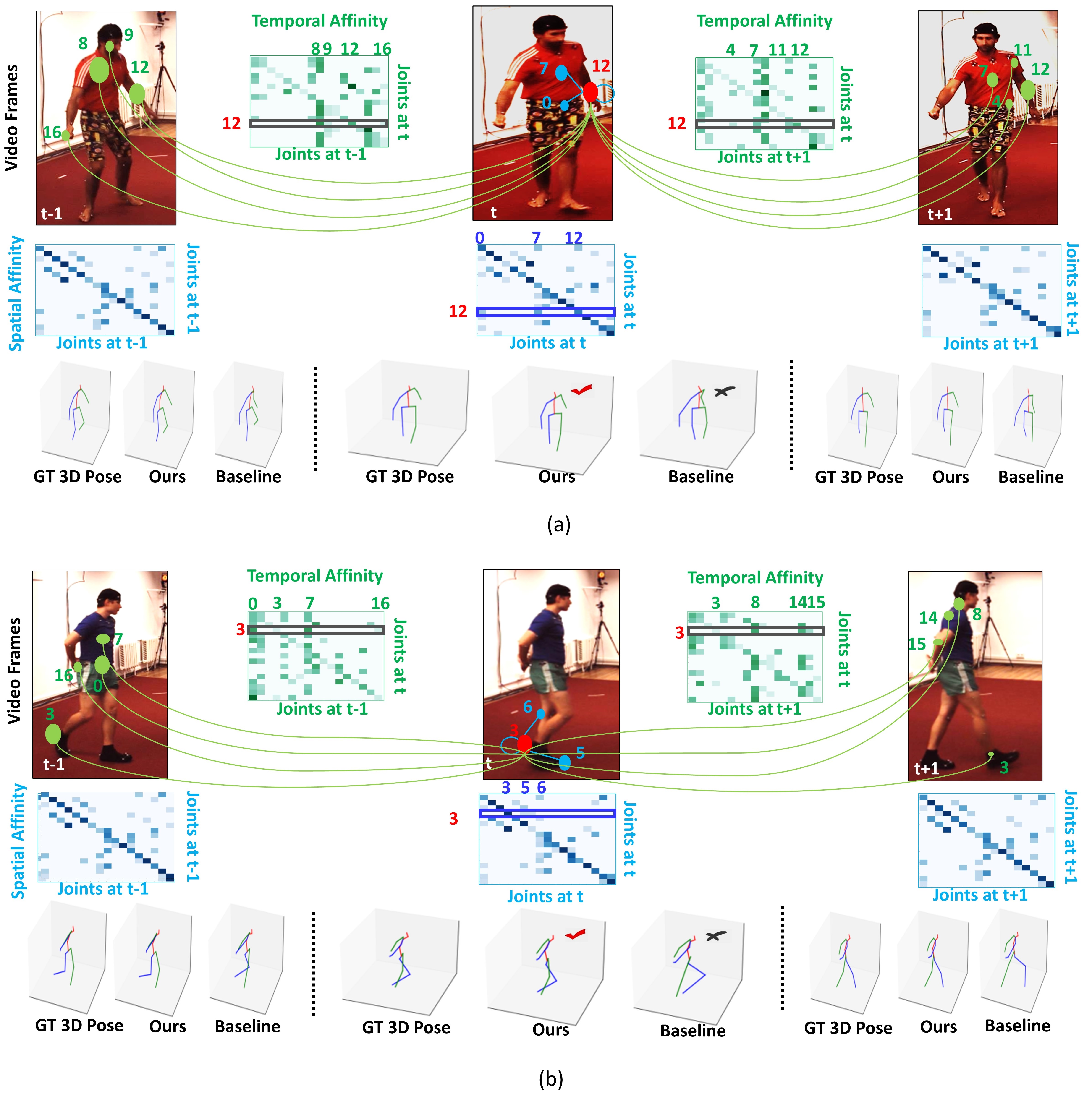}
\caption{Visualization.
We show the estimated 3D pose in a \textit{WalkingDog} video (a) and a \textit{Walking} video (b).
As expected,
spatial and temporal affinity matrices are dynamically adjusted,
which allows DG-Net to reduce ambiguity by learning richer pose relations.
}
\label{VC3}
\vspace{-0.5cm}
\end{figure*}

\begin{figure*}[t]
\centering
\includegraphics[width=0.98\textwidth]{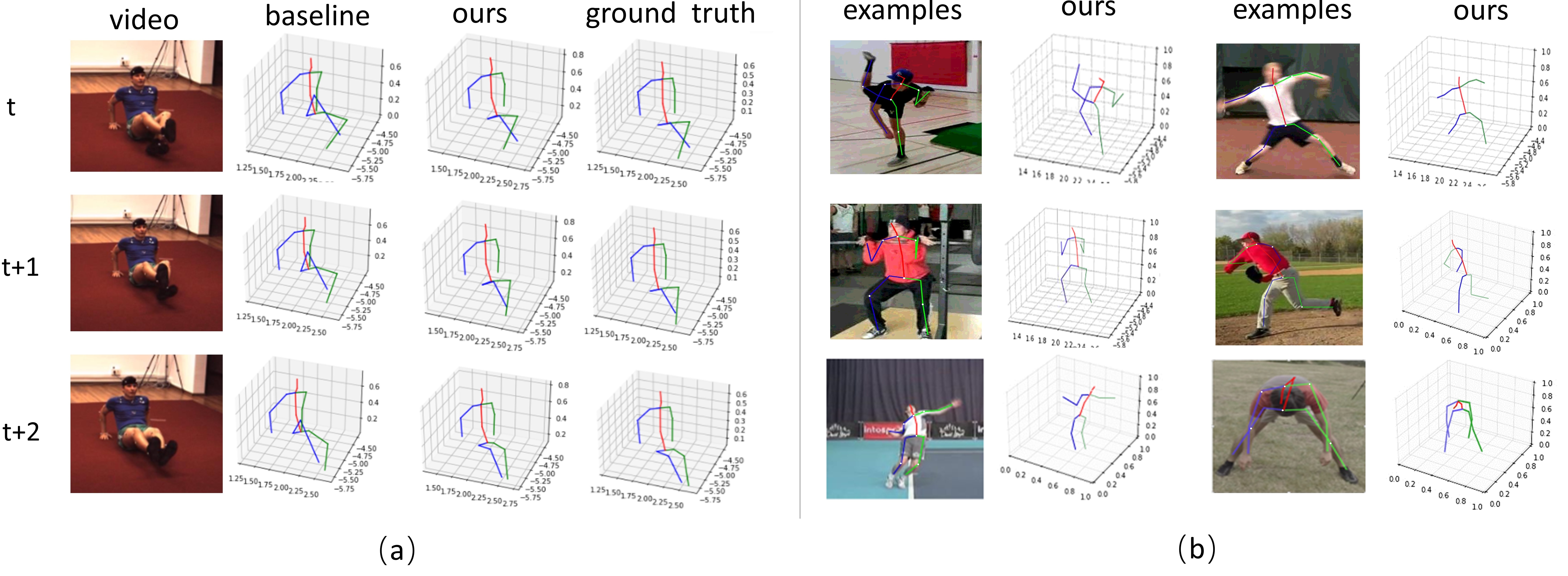}
\caption{Visualization. (a) shows the results in Human3.6M\cite{zhang2013actemes} dataset; (b) shows the results in MPI-INF-3DHP \cite{mono_3dhp2017}  and Upenn Action \cite{zhang2013actemes} datasets.
}
\label{VC33}
\vspace{-0.5cm}
\end{figure*}

\textbf{Blocks \& Frames}.
In Table \ref{tab:blockframe},
the performance is the best with 5 DG-Conv blocks.
Additionally,
the 4-frame setting is comparable with the 5-frame setting.
To reduce computation cost,
we choose the 4-frame setting.

\textbf{Robustness-to-Noise}.
We evaluate the robustness of our approach (T=4) to corrupted 2D joint locations.
In this experiment,
all approaches are trained using the ground truth 2D and 3D pairs in Human3.6M dataset.
Then we test on inputs corrupted by different levels of Gaussian noise.
As shown in Table~\ref{noise},
our method obtains the best result and maintains robustness to all levels of Gaussian noise,
due to our dynamical affinity of joint connections.

\textbf{Cross Actions Generalization Ability}.
We train our model on one of the 15 actions in the Human3.6M \cite{h36m_pami} dataset and test on all actions.
As shown in Figure \ref{VC333},
our DG-Net significantly improves the performance compared with baseline and outperforms the SOTA video 2D-pose lifter \cite{Cai_2019_ICCV} by a large margin.
It demonstrates that our DG-Net is capable of adjusting human-joint affinity dynamically to enhance 3D pose estimation,
depending on the personalized poses of different actions in the video.

\textbf{Other Detailed Designs.} We further investigate other detailed designs  in Table \ref{tab:detail}.
First,
we evaluate our multi-level supervision.
As expected,
the total loss in Eq. (\ref{alllevel}) leads to a better performance,
compared to the network loss in Eq. (\ref{networklevel}).
It illustrates that,
adding 3D supervision in each block can effectively regularize our DG-Net with more discriminative pose representations.
Second,
we follow the good practice in \cite{semanticsgcn} via embedding spatial nonlocal operations in our DG-Conv block.
As expected,
it achieves a better performance by learning global pose relations.
Third,
we evaluate the impact of coordinate operations.
In general,
there are two coordinate systems for 3D human pose.
The first uses the root-relative 3D joint coordinates in the camera coordinate system.
We denote it as $P_{c}=(X_c,Y_c,Z_c)$.
The second refers to the concatenation of
the pixel UV coordinates extracted from 2D pose detector
and
the predicted depth of each joint.
We denote it as $P_{p}=(U,V,D)$.
The non-balanced setting in Table XI refers to the prediction of $P_{c}$.
For the balanced setting,
we use the camera intrinsic parameters ($c_{x}$, $c_{y}$, $f$) to project UV coordinates into camera coordinate system,
where
$\overline{X_{c}}=Z_{c}\frac{U-c_{x}}{f}$ and $ \overline{Y_{c}}=Z_{c}\frac{V-c_{y}}{f}$.
Then,
we compute the final prediction of 3D pose via average balancing,
where
$P=(\frac{X_{c}+\overline{X_{c}}}{2},\frac{Y_{c}+\overline{Y_{c}}}{2},Z_{c})$.
As expected,
such balance takes advantage of both systems to achieve a better 3D estimation.
Fourth,
we evaluate the impact of 2D pose estimators.
With a more accurate 2D estimator \cite{cpn},
the result is better.
Finally,
we perform both the unified form and factorized form of our DG-Conv block,
i.e.,
the unified form refers to the spatio-temporal convolution,
while
the factorized form (i.e., our design) refers to the spatial convolution and temporal convolution in our DG-Conv.
We can see that,
the MPJPE of unified/factorized forms are 48.9/45.3 mm,
which shows the effectiveness of our proposed architecture.

\subsection{Visualization}
We visualize DG-Net via the estimated 3D pose in a $WalkingDog$ video and a $Walking$ video.
Moreover,
we show spatial affinity (per frame) in DSG,
and temporal affinity (between two frames) in DTG.
All these affinity matrices are extracted from the 1st DG-Conv block.
As shown in Fig. \ref{VC3},
spatial/temporal human-joint affinity matrices change over time.
This allows DG-Net to dynamically exploit pose context for each joint to boost 3D estimation,
e.g.,
in  Fig. \ref{VC3} (a),  for Joint12 (left wrist) at $t$,
spatial affinity discovers Joint0 (hip) and Joint7 (spine) as extra spatial context at $t$,
while
temporal affinity discovers Joint4 (left hip), Joint7 (spine), Joint11 (left shoulder) as extra temporal context at $t+1$,
and Joint8 (thorax), Joint9 (nose), Joint16 (right wrist) as extra temporal context at $t-1$.
Those joints can effectively tell that,
Joint12 is straighten and moving similarly with the upper body.
As a result,
our DG-Net correctly estimate Joint12 at $t$.
On the contrary,
baseline distorts this joint with a wrong depth,
due to the fixed spatial and temporal graph convolutions.
Additional qualitative results on Human3.6M\cite{h36m_pami}, MPI-INF-3DHP \cite{mono_3dhp2017}  and Upenn Action \cite{zhang2013actemes} are shown in Figure \ref{VC33}. Our model also achieves good results in outdoor scenes and in the wild, which proves the generalization of our model.

\section{Conclusion}
In this work,
we propose a novel DG-Net for video 3D pose estimation.
By learning human-joint affinity dynamically,
it can build up effective joint relations to reduce spatial and temporal ambiguity caused by complex pose variations in videos.
Extensive experiments demonstrate high accuracy and effectiveness of our DG-Net.


\ifCLASSOPTIONcaptionsoff
  \newpage
\fi


\bibliographystyle{IEEEtran}

\begin{thebibliography}{10}
\providecommand{\url}[1]{#1}
\csname url@samestyle\endcsname
\providecommand{\newblock}{\relax}
\providecommand{\bibinfo}[2]{#2}
\providecommand{\BIBentrySTDinterwordspacing}{\spaceskip=0pt\relax}
\providecommand{\BIBentryALTinterwordstretchfactor}{4}
\providecommand{\BIBentryALTinterwordspacing}{\spaceskip=\fontdimen2\font plus
\BIBentryALTinterwordstretchfactor\fontdimen3\font minus
  \fontdimen4\font\relax}
\providecommand{\BIBforeignlanguage}[2]{{%
\expandafter\ifx\csname l@#1\endcsname\relax
\typeout{** WARNING: IEEEtran.bst: No hyphenation pattern has been}%
\typeout{** loaded for the language `#1'. Using the pattern for}%
\typeout{** the default language instead.}%
\else
\language=\csname l@#1\endcsname
\fi
#2}}
\providecommand{\BIBdecl}{\relax}
\BIBdecl

\bibitem{Lin_2017_CVPR}
M.~Lin, L.~Lin, X.~Liang, K.~Wang, and H.~Cheng, ``Recurrent 3d pose sequence
  machines,'' in \emph{CVPR}, July 2017.

\bibitem{eccv2018temporal}
M.~R.~I. Hossain and J.~Little, ``Exploiting temporal information for 3d human
  pose estimation,'' \emph{ECCV}, 2018.

\bibitem{pavllo:videopose3d:2019}
D.~Pavllo, C.~Feichtenhofer, D.~Grangier, and M.~Auli, ``3d human pose
  estimation in video with temporal convolutions and semi-supervised
  training,'' in \emph{CVPR}, 2019.

\bibitem{semanticsgcn}
L.~Zhao, X.~Peng, Y.~Tian, M.~Kapadia, and D.~N. Metaxas, ``Semantic graph
  convolutional networks for 3d human pose regression,'' \emph{CVPR}, 2019.

\bibitem{Ci_2019_ICCV}
H.~Ci, C.~Wang, X.~Ma, and Y.~Wang, ``Optimizing network structure for 3d human
  pose estimation,'' in \emph{ICCV}, October 2019.

\bibitem{Cai_2019_ICCV}
Y.~Cai, L.~Ge, J.~Liu, J.~Cai, T.-J. Cham, J.~Yuan, and N.~M. Thalmann,
  ``Exploiting spatial-temporal relationships for 3d pose estimation via graph
  convolutional networks,'' in \emph{ICCV}, October 2019.

\bibitem{dantone2013human}
M.~Dantone, J.~Gall, C.~Leistner, and L.~Van~Gool, ``Human pose estimation
  using body parts dependent joint regressors,'' in \emph{Proceedings of the
  IEEE Conference on Computer Vision and Pattern Recognition}, 2013, pp.
  3041--3048.

\bibitem{wei2016convolutional}
S.-E. Wei, V.~Ramakrishna, T.~Kanade, and Y.~Sheikh, ``Convolutional pose
  machines,'' in \emph{Proceedings of the IEEE conference on Computer Vision
  and Pattern Recognition}, 2016, pp. 4724--4732.

\bibitem{toshev2014deeppose}
A.~Toshev and C.~Szegedy, ``Deeppose: Human pose estimation via deep neural
  networks,'' in \emph{Proceedings of the IEEE conference on computer vision
  and pattern recognition}, 2014, pp. 1653--1660.

\bibitem{cpn}
Y.~Chen, Z.~Wang, Y.~Peng, Z.~Zhang, G.~Yu, and J.~Sun, ``Cascaded pyramid
  network for multi-person pose estimation,'' in \emph{CVPR}, 2018.

\bibitem{hourglass}
A.~Newell, K.~Yang, and J.~Deng, ``Stacked hourglass networks for human pose
  estimation.'' in \emph{ECCV}, 2016.

\bibitem{qiu2020dgcn}
Z.~Qiu, K.~Qiu, J.~Fu, and D.~Fu, ``Dgcn: Dynamic graph convolutional network
  for efficient multi-person pose estimation.'' in \emph{AAAI}, 2020, pp.
  11\,924--11\,931.

\bibitem{mono_3dhp2017}
D.~Mehta, H.~Rhodin, D.~Casas, P.~Fua, O.~Sotnychenko, W.~Xu, and C.~Theobalt,
  ``Monocular 3d human pose estimation in the wild using improved cnn
  supervision,'' in \emph{3DV}, 2017.

\bibitem{volumetric}
G.~Pavlakos, X.~Zhou, K.~G. Derpanis, and K.~Daniilidis, ``Coarse-to-fine
  volumetric prediction for single-image 3{D} human pose,'' in \emph{CVPR},
  2017.

\bibitem{integral}
X.~Sun, B.~Xiao, F.~Wei, S.~Liang, and Y.~Wei, ``Integral human pose
  regression,'' in \emph{ECCV}, 2018.

\bibitem{ordinal}
G.~Pavlakos, X.~Zhou, and K.~Daniilidis, ``Ordinal depth supervision for 3d
  human pose estimation,'' in \emph{CVPR}, 2018.

\bibitem{simple}
J.~Martinez, R.~Hossain, J.~Romero, and J.~J. Little, ``A simple yet effective
  baseline for 3d human pose estimation,'' in \emph{ICCV}, 2017.

\bibitem{lin2019trajectory}
J.~Lin and G.~H. Lee, ``Trajectory space factorization for deep video-based 3d
  human pose estimation,'' \emph{arXiv preprint arXiv:1908.08289}, 2019.

\bibitem{wang2020motion}
J.~Wang, S.~Yan, Y.~Xiong, and D.~Lin, ``Motion guided 3d pose estimation from
  videos,'' \emph{arXiv preprint arXiv:2004.13985}, 2020.

\bibitem{alej2016stacked}
A.~Newell, K.~Yang, and J.~Deng, ``Stacked hourglass networks for human pose
  estimation,'' 2016.

\bibitem{distance}
F.~Moreno-Noguer, ``3d human pose estimation from a single image via distance
  matrix regression,'' in \emph{CVPR}, 2017.

\bibitem{Lee_2018_ECCV}
K.~Lee, I.~Lee, and S.~Lee, ``Propagating lstm: 3d pose estimation based on
  joint interdependency,'' in \emph{ECCV}, September 2018.

\bibitem{HenaffarXiv2015}
M.~Henaff, J.~Bruna, and Y.~LeCun, ``Deep convolutional networks on
  graph-structured data,'' in \emph{arXiv}, 2015.

\bibitem{LiICLR2016}
Y.~Li, D.~Tarlow, M.~Brockschmidt, and R.~Zemel, ``Gated graph sequence neural
  networks,'' in \emph{ICLR}, 2016.

\bibitem{Bronstein2017}
M.~M. Bronstein, J.~Bruna, Y.~LeCun, A.~Szlam, and P.~Vandergheynst,
  ``Geometric deep learning: going beyond euclidean data,'' \emph{IEEE Signal
  Processing Magazine}, 2017.

\bibitem{maxgcn}
T.~N. Kipf and M.~Welling, ``Semi-supervised classification with graph
  convolutional networks,'' \emph{ICLR}, 2017.

\bibitem{PetarICLR18}
P.~Veli\v{c}kovi\'{c}, G.~Cucurull, A.~Casanova, A.~Romero, P.~Li\`{o}, and
  Y.~Bengio, ``Graph attention networks,'' in \emph{ICLR}, 2018.

\bibitem{stgcn}
S.~Yan, Y.~Xiong, and D.~Lin, ``Spatial temporal graph convolutional networks
  for skeleton-based action recognition,'' \emph{AAAI}, 2018.

\bibitem{wu2019learning}
J.~Wu, L.~Wang, L.~Wang, J.~Guo, and G.~Wu, ``Learning actor relation graphs
  for group activity recognition,'' in \emph{Proceedings of the IEEE/CVF
  Conference on Computer Vision and Pattern Recognition}, 2019, pp. 9964--9974.

\bibitem{dynamicgraph_gesture}
Y.~Chen, L.~Zhao, X.~Peng, J.~Yuan, and D.~N. Metaxas, ``Construct dynamic
  graphs for hand gesture recognition via spatial-temporal attention,''
  \emph{BMVC}, 2019.

\bibitem{Shi_2019_CVPR}
L.~Shi, Y.~Zhang, J.~Cheng, and H.~Lu, ``Two-stream adaptive graph
  convolutional networks for skeleton-based action recognition,'' in
  \emph{Proceedings of the IEEE/CVF Conference on Computer Vision and Pattern
  Recognition (CVPR)}, June 2019.

\bibitem{tran2018closer}
D.~Tran, H.~Wang, L.~Torresani, J.~Ray, Y.~LeCun, and M.~Paluri, ``A closer
  look at spatiotemporal convolutions for action recognition,'' in \emph{CVPR},
  2018, pp. 6450--6459.

\bibitem{rnnpose}
H.-S. Fang, Y.~Xu, W.~Wang, X.~Liu, and S.-C. Zhu, ``Learning pose grammar to
  encode human body configuration for 3d pose estimation,'' in \emph{AAAI},
  2018.

\bibitem{adversialpose}
W.~Yang, W.~Ouyang, X.~Wang, J.~Ren, H.~Li, and X.~Wang, ``3d human pose
  estimation in the wild by adversarial learning,'' in \emph{CVPR}, 2018.

\bibitem{Wang_2019_ICCV}
J.~Wang, S.~Huang, X.~Wang, and D.~Tao, ``Not all parts are created equal: 3d
  pose estimation by modeling bi-directional dependencies of body parts,'' in
  \emph{ICCV}, October 2019.

\bibitem{Xu_2020_CVPR}
J.~Xu, Z.~Yu, B.~Ni, J.~Yang, X.~Yang, and W.~Zhang, ``Deep kinematics analysis
  for monocular 3d human pose estimation,'' in \emph{CVPR}, June 2020.

\bibitem{liu2020gast}
J.~Liu, Y.~Guang, and J.~Rojas, ``Gast-net: Graph attention spatio-temporal
  convolutional networks for 3d human pose estimation in video,'' \emph{arXiv
  preprint arXiv:2003.14179}, 2020.

\bibitem{Liu_2020_CVPR}
R.~Liu, J.~Shen, H.~Wang, C.~Chen, S.-c. Cheung, and V.~Asari, ``Attention
  mechanism exploits temporal contexts: Real-time 3d human pose
  reconstruction,'' in \emph{CVPR}, June 2020.

\bibitem{Yasin_2016_CVPR}
H.~Yasin, U.~Iqbal, B.~Kruger, W.~andreas, and J.~Gall, ``A dual-source
  approach for 3d pose estimation from a single image,'' in \emph{CVPR}, June
  2016.

\bibitem{VNect_SIGGRAPH2017}
\BIBentryALTinterwordspacing
D.~Mehta, S.~Sridhar, O.~Sotnychenko, H.~Rhodin, M.~Shafiei, H.-P. Seidel,
  W.~Xu, D.~Casas, and C.~Theobalt, ``Vnect: Real-time 3d human pose estimation
  with a single rgb camera,'' vol.~36, no.~4, 2017. [Online]. Available:
  \url{http://gvv.mpi-inf.mpg.de/projects/VNect/}
\BIBentrySTDinterwordspacing

\bibitem{h36m_pami}
C.~Ionescu, D.~Papava, V.~Olaru, and C.~Sminchisescu, ``Human3.6m: Large scale
  datasets and predictive methods for 3d human sensing in natural
  environments,'' \emph{PAMI}, 2014.

\bibitem{humaneva}
L.~Sigal, A.~O. Balan, and M.~J. Black, ``Humaneva: Synchronizedvideo and
  motion capture dataset and baseline algorithm forevaluation of articulated
  human motion,'' in \emph{IJCV}, 2010.

\bibitem{maskrcnn}
K.~He, G.~Gkioxari, P.~Dollár, and R.~Girshick, ``Mask r-cnn,'' \emph{ICCV},
  2017.

\bibitem{nonlocal}
X.~Wang, R.~Girshick, A.~Gupta, and K.~He, ``Non-local neural networks,''
  \emph{CVPR}, 2018.

\bibitem{zhang2013actemes}
W.~Zhang, M.~Zhu, and K.~G. Derpanis, ``From actemes to action: A
  strongly-supervised representation for detailed action understanding,'' in
  \emph{ICCV}, 2013, pp. 2248--2255.

\end{thebibliography}

\end{document}